\documentclass[runningheads]{llncs}

\usepackage{subcaption}
\usepackage[]{microtype}
\UseMicrotypeSet[protrusion]{basicmath} 
\usepackage{lmodern}
\usepackage{amssymb,amsmath}
\usepackage[T1]{fontenc}
\usepackage[utf8]{inputenc}
\usepackage{textcomp} 
\usepackage{graphicx}
\usepackage{mdframed}
\usepackage{import}
\usepackage{todonotes}
\usepackage{fancyvrb}
\usepackage{listings}
\usepackage{import}
\usepackage{todonotes}
\usepackage{wrapfig}
\usepackage{nicefrac}
\usepackage{ragged2e}
\usepackage{caption}
\usepackage{mystyle}

\usepackage{xcolor}
 
\definecolor{codegreen}{rgb}{0,0.6,0}
\definecolor{codegray}{rgb}{0.5,0.5,0.5}
\definecolor{codepurple}{rgb}{0.58,0,0.82}
\definecolor{backcolour}{rgb}{0.95,0.95,0.92}

\lstdefinestyle{mystyle}{
    backgroundcolor=\color{backcolour},   
    commentstyle=\color{codegreen},
    keywordstyle=\color{magenta},
    numberstyle=\tiny\color{codegray},
    stringstyle=\color{codepurple},
    basicstyle=\ttfamily\footnotesize,
    breakatwhitespace=false,         
    breaklines=true,                 
    captionpos=b,                    
    keepspaces=true,                 
    numbers=left,                    
    numbersep=5pt,                  
    showspaces=false,                
    showstringspaces=false,
    showtabs=false,                  
    tabsize=2
}
 
\newmdtheoremenv[style=2]{mdexample}{Example}
\newmdtheoremenv[style=2]{mdobs}{Observation}
\newmdtheoremenv[style=2, nobreak=true]{mddef}{Definition}
\newmdtheoremenv[style=2, nobreak=true]{mdtheo}{Theorem}
\newmdtheoremenv[style=2]{mdlem}{Theorem}
\newmdtheoremenv[style=2, nobreak=true]{mdcorr}{Corollary}
\newmdtheoremenv[style=2]{mdprop}{Proposition}
\newmdtheoremenv[style=2]{mdprune}{Pruning Rule}
\newmdtheoremenv[style=proofstyle]{mdproof}{Proof}

\usepackage{amsmath,amsfonts,bm}

\def\figref#1{figure~\ref{#1}}

\def\secref#1{section~\ref{#1}}

\def\eqref#1{equation~\ref{#1}}

\def\ceil#1{\lceil #1 \rceil}

\def\1{\bm{1}}

\DeclareMathAlphabet{\mathsfit}{\encodingdefault}{\sfdefault}{m}{sl}
\SetMathAlphabet{\mathsfit}{bold}{\encodingdefault}{\sfdefault}{bx}{n}

\newcommand{\E}{\mathop{\mathbb{E}}}

\newcommand{\softmax}{\mathrm{softmax}}

\DeclareMathOperator*{\argmax}{arg\,max}

\newcommand{\Demos}{X}

\newcommand{\setof}[1]{\ensuremath{\left\{#1\right\}}}

\newcommand{\f}{\varphi}
\newcommand{\boolcube}[1]{\ensuremath{\setof{0, 1}^{#1}}}
\newcommand{\eqdef}{\mathrel{\stackrel{\makebox[0pt]{\mbox{\normalfont\tiny def}}}{=}}}

\newcommand{\Reals}{\mathbb{R}}
\newcommand{\PosReals}{\mathbb{R}_{\ge 0}}

\newcommand{\Nat}{\mathbb{N}}

\newcommand{\Prob}{\Pr}

\newcommand{\true}{\mathit{true}}
\newcommand{\false}{\mathit{false}}

\renewcommand{\eqref}[1]{(\ref{eq:#1})}

\renewcommand{\secref}[1]{Sec.~\ref{sec:#1}}
\renewcommand{\figref}[1]{Fig.~\ref{fig:#1}}

\newcommand{\thmref}[1]{Theorem~\ref{thm:#1}}

\usepackage[firstpage]{draftwatermark}\SetWatermarkText{\hspace*{5.5in}\raisebox{9in}{\includegraphics[scale=0.1]{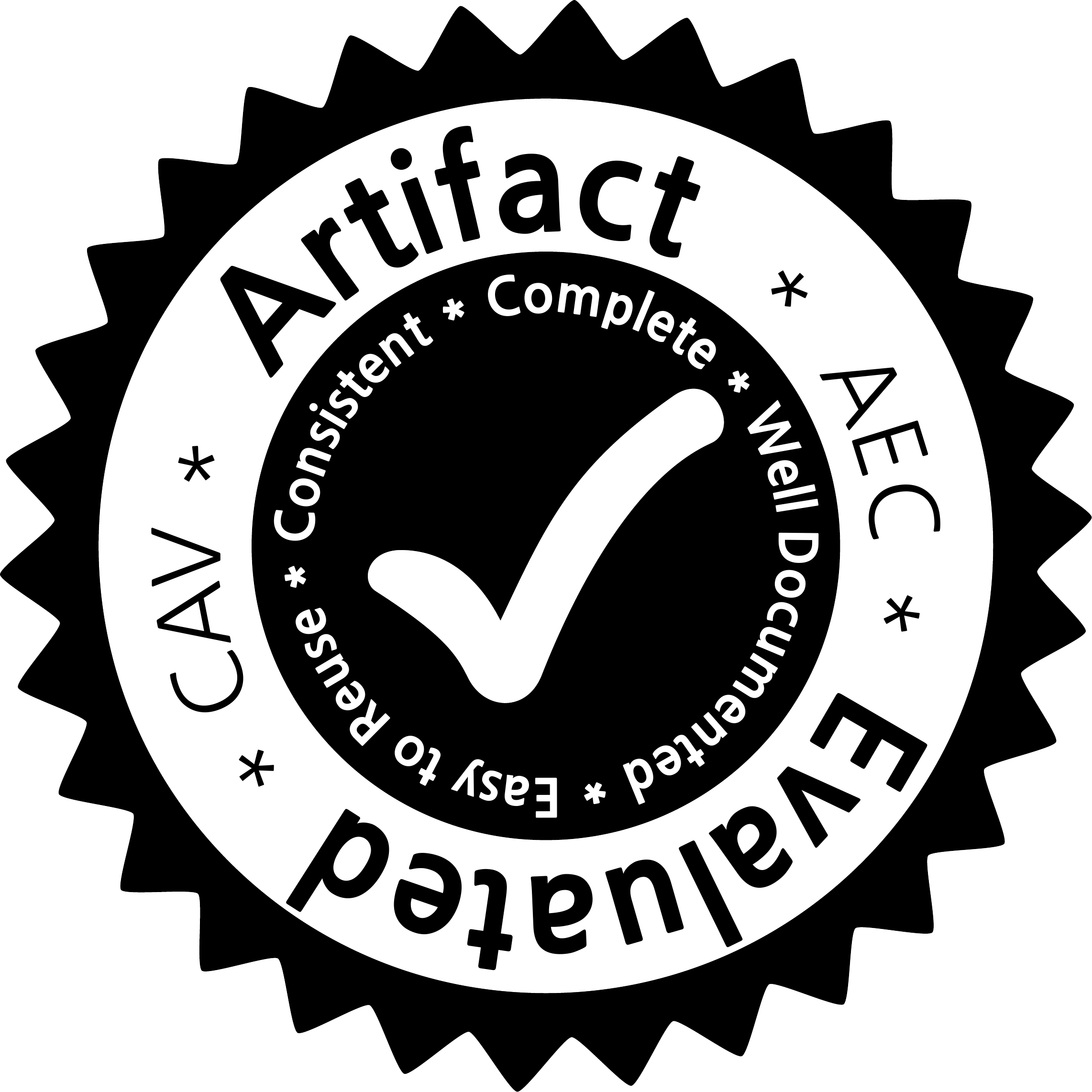}}}\SetWatermarkAngle{0}
\begin{document}
\title{ Maximum Causal Entropy Specification Inference from
  Demonstrations \vspace{-1em} }
%
%
\author{Marcell Vazquez-Chanlatte\and Sanjit A. Seshia}
%
\authorrunning{Vazquez-Chanlatte et al.}
%
\institute{University of California, Berkeley, USA}
%
\maketitle 
\vspace{-2em}
\begin{abstract} 
  In many settings, such as robotics, demonstrations provide a natural
  way to specify tasks. However, most methods for learning from
  demonstrations either do not provide guarantees that the learned
  artifacts can be safely composed or do not explicitly capture
  temporal properties. Motivated by this deficit, recent works have
  proposed learning Boolean \emph{task specifications}, a class of
  Boolean non-Markovian rewards which admit well-defined composition
  and explicitly handle historical dependencies. This work continues
  this line of research by adapting maximum \emph{causal} entropy
  inverse reinforcement learning to estimate the posteriori
  probability of a specification given a multi-set of
  demonstrations. The key algorithmic insight is to leverage the
  extensive literature and tooling on reduced ordered binary decision
  diagrams to efficiently encode a time unrolled Markov Decision
  Process. This enables transforming a na\"ive algorithm with running
  time exponential in the episode length, into a polynomial time
  algorithm.
\end{abstract}

\raggedbottom

\section{Introduction}\label{sec:intro}

\begin{wrapfigure}[10]{r}{4cm}  
  \centering
  \vspace{-2em}
  \scalebox{0.6}{ 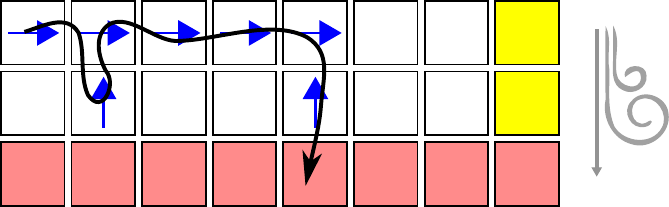 }
  \caption{Example of an agent unsuccessfully demonstrating the task
    ``reach a yellow tile while avoiding red tiles''
    .\label{fig:enterlava2}}
\end{wrapfigure}
In many settings, episodic demonstrations provide a natural and robust
mechanism to partially specify a task, even in the presence of
errors. For example, consider the agent operating in the gridworld
illustrated in~\figref{enterlava2}. Blue arrows denote intended
actions and the solid black arrow shows the agent's actual path.
This path can stochastically differ from the blue arrows due to a
downward wind. One might naturally ask: ``What task was this agent
attempting to perform?'' Even without knowing if this was a positive
or negative example, based on the agent's state/action sequence, one
can reasonably infer the agent's intent, namely, ``reach the yellow
tile while avoiding the red tiles.'' Compared with traditional
learning from positive and negative examples, this is somewhat
surprising, particularly given that the task is never actually
demonstrated in~\figref{enterlava2}.

This problem, inferring intent from demonstrations, has received a
fair amount of attention over the past two decades particularly within
the robotics
community~\cite{ng2000algorithms}\cite{ziebart2008maximum}\cite{finn2016guided}\cite{max_ent_spec_learning}.  In this
literature, one traditionally models the demonstrator as operating
within a dynamical system whose transition relation only depends on
the current state and action (called the Markov condition). However,
even if the dynamics are Markovian, many tasks are naturally modeled
in history dependent (non-Markovian) terms, e.g., ``if the robot
enters a blue tile, then it must touch a brown tile \emph{before}
touching a yellow tile''. Unfortunately, most methods for learning
from demonstrations either do not provide guarantees that the learned
artifacts (e.g. rewards) can be safely composed or do not explicitly
capture history dependencies~\cite{max_ent_spec_learning}.
  
Motivated by this deficit, recent works have proposed specializing to
\textbf{task specifications}, a class of Boolean non-Markovian rewards
induced by formal languages. This additional structure admits
well-defined compositions and explicitly captures temporal
dependencies~\cite{kasenberg2017interpretable}\cite{max_ent_spec_learning}.
A particularly promising direction has been to adapt maximum entropy
inverse reinforcement learning~\cite{ziebart2008maximum} to task
specifications, enabling a form of robust specification inference,
even in the presence unlabeled demonstration
errors~\cite{max_ent_spec_learning}.

However, while powerful, the principle of maximum entropy is limited
to settings where the dynamics are deterministic or agents that use
open-loop policies~\cite{ziebart2008maximum}. This is because the
principle of maximum entropy incorrectly allows the agent's predicted
policy to depend on future state values resulting in an overly
optimistic agent~\cite{DBLP:journals/corr/abs-1805-00909}.  For
instance, in our gridworld example (\figref{enterlava2}), the
principle of maximum entropy would discount the possibility of
slipping, and thus we would not forecast the agent to correct its
trajectory after slipping once.

This work continues this line of research by instead using the
principle of maximum \emph{causal} entropy, which generalizes the
principle of maximum entropy to general stochastic decision
processes~\cite{mce}. While a conceptually straightforward extension,
a na\"ive application of maximum \emph{causal} entropy inverse
reinforcement learning to non-Markovian rewards results in an
algorithm with run-time exponential in the episode length, a
phenomenon sometimes known as the \textbf{curse of
history}~\cite{pineau2003point}.  The key algorithmic insight in this
paper is to leverage the extensive literature and tooling on Reduced
Ordered Binary Decision Diagrams (BDDs)~\cite{bryant1992symbolic} to efficiently encode the
time unrolled composition of the dynamics and task specification.
This allows us to translate a na\"ive exponential time algorithm
into a polynomial time algorithm. In particular, we shall show
that this BDD has size at most linear in the episode
length making inference comparatively efficient.

\subsection{Related Work:}

Our work is intimately related to the fields of Inverse Reinforcement
Learning and Grammatical Inference.  \textbf{Grammatical
inference}~\cite{de2010grammatical} refers to the well-developed
literature on learning a formal grammar (often an automaton) from data.
Examples include learning the smallest automata that in consistent
with a set of positive and negative
strings~\cite{heule2010exact}\cite{de2010grammatical} or learning an
automaton using membership and equivalence
queries~\cite{angluin1987learning}. This and related work can be seen
as extending these methods to unlabeled and potentially noisy
demonstrations, where demonstrations differ from examples due to the
existence of a dynamics model. This notion of demonstration derives
from the Inverse Reinforcement Learning literature.

In \textbf{Inverse Reinforcement Learning}
(IRL)~\cite{ng2000algorithms} the demonstrator, operating in a
stochastic environment, is assumed to attempt to (approximately)
optimize some unknown reward function over the trajectories. In
particular, one traditionally assumes a trajectory's reward is the sum
of state rewards of the trajectory. This formalism offers a succinct
mechanism to encode and generalize the goals of the demonstrator to
new and unseen environments.

In the IRL framework, the problem of learning from demonstrations can
then be cast as a Bayesian inference
problem~\cite{ramachandran2007bayesian} to predict the most probable
reward function. To make this inference procedure well-defined and
robust to demonstration/modeling noise, Maximum
Entropy~\cite{ziebart2008maximum} and Maximum Causal
Entropy~\cite{mce} IRL appeal to the principles of maximum
entropy~\cite{jaynes1957information} and maximum causal entropy
respectively~\cite{mce}. This results in a likelihood over the
demonstrations which is no more committed to any particular behavior
than what is required to match observed statistical features, e.g.,
average distance to an obstacle.  While this approach was initially
limited to rewards represented as linear combinations of scalar
features, IRL has been successfully adapted to arbitrary function
approximators such as Gaussian processes~\cite{NIPS2011_4420} and
neural networks~\cite{finn2016guided}. As stated in the introduction,
while powerful, traditional IRL provides no principled mechanism for
composing the resulting rewards.

\textbf{Compositional RL:} To address this deficit, composition using
soft optimality has recently received a fair amount of attention;
however, the compositions are limited to either strict disjunction (do
X \emph{or} Y)~\cite{todorov2007linearly}~\cite{todorov2008general} or
conjunction (do X \emph{and}
Y)~\cite{haarnoja2018composable}. Further, this soft optimality only
bounds the deviation from simultaneously optimizing both
rewards. Thus, optimizing the composition does not preclude violating
safety constraints embedded in the rewards (e.g., do not enter the red
tiles).

\textbf{Logic based IRL:} Another promising approach for introducing
composationality has been the recent research on automata and logic
based encodings of rewards
\cite{icarte2018using}\cite{jothimurugan2019composable} which admit well defined compositions. To this end, work has been done on
inferring Linear Temporal Logic (LTL) formulas by finding the
specification that minimizes the expected number of violations by an
optimal agent compared to the expected number of violations by an
agent applying actions uniformly at
random~\cite{kasenberg2017interpretable}. The computation of the
optimal agent's expected violations is done via dynamic programming on
the explicit product of the deterministic Rabin
automaton~\cite{farwer2002omega} of the specification and the state
dynamics. A fundamental drawback of this procedure is that due to the
curse of history, it incurs a heavy run-time cost, even on simple two
state and two action Markov Decision Processes. Additionally, as with
early work on grammatical inference and IRL, these techniques do not
produce likelihood estimates amenable to Bayesian inference.

\textbf{Maximum Entropy Specification Inference:}
In our previous work~\cite{max_ent_spec_learning}, we adapted maximum
entropy IRL to learn task specifications. Similar to standard maximum
entropy IRL, this technique produces robust likelihood
estimates. However, due to the use of the principle of maximum
entropy, rather than maximum \emph{causal} entropy, this model is
limited to settings where the dynamics are deterministic or agents
with open-loop policies~\cite{ziebart2008maximum}.

\textbf{Inference using BDDs:} This work makes heavy use of Binary
Decision Diagrams (BDDs)~\cite{bryant1992symbolic} which are
frequently used in symbolic value iteration for Markov Decision
Processes~\cite{hoey1999spudd} and reachability analysis for
probabilistic systems\cite{KNP11}. However, the literature has largely
relied on Multi-Terminal BDDs to encode the transition probabilities
for a \textbf{single} time step. In contrast, this work introduces a
two-terminal encoding based on the finite unrolling of a probabilistic
circuit. To the best of our knowledge, the most similar usage of BDDs
for inference appears in the independently discovered literal weight
based encoding of~\cite{Holtzen19} - although their encoding does not
directly support non-determinism or state-indexed random variables.

\textbf{Contributions:}
The primary contributions of this work are two fold. First, we
leverage the principle of maximum causal entropy to provide the
likelihood of a specification given a set of demonstrations. This
formulation removes the deterministic and/or open-loop restriction
imposed by prior work based on the principle of maximum entropy.
Second, to mitigate the curse of history, we propose using a
BDD to encode the time unrolled Markov Decision Process
that the maximum causal entropy forecaster is defined over. We prove that this BDD
has size that grows linearly with the horizon and quasi-linearly with
the number of actions.  Furthermore, we prove that our derived
likelihood estimates are robust to the particular reward associated
with satisfying the specification.  Finally, we provide an initial
experimental validation of our method. An overview of this pipeline is
provided in~\figref{overview}.

\vspace{-1.2em}
\section{Problem Setup}\label{sec:background}
\vspace{-0.6em}

We seek to learn task specifications from demonstrations provided by a
teacher who executes a sequence of actions that probabilistically
change the system state. For simplicity, we assume that the set of
actions and states are finite and fully observed. Further,
until~\secref{discount}, we shall assume that all demonstrations
are a fixed length, $\tau \in \Nat$. Formally, we begin by
modeling the underlying dynamics as a probabilistic automaton.

\begin{mddef}
  \begin{flushleft}
    A \textbf{probabilistic automaton} (PA) is a tuple
    $(S, s_0, A, \delta) $, where $ S $ is the finite set of states,
    $s_0 \in S$ is the initial state, $ A $ is a finite set of
    actions, and $ \delta$ specifies the transition probability of
    going from state $ s $ to state $ s' $ given action $ a $,
    i.e. $\delta(s, a, s') = \Prob(s'~|~s,a) $.

    \vspace{0.6em} A \textbf{trace}\footnote{sometimes referred to as a
      trajectory or behavior.}, $\xi$, is a sequence of (action,
    state) pairs implicitly starting from $s_0$.  A trace of
    length $\tau \in \Nat$ is an element of $(A\times S)^{\tau}$.
  \end{flushleft}
\end{mddef}

\begin{figure}[h]
  \begin{subfigure}[t]{0.45\linewidth}
    \begin{center}
      \scalebox{1.3}{ 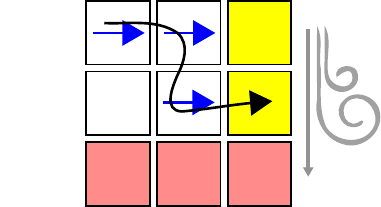 }
      \caption{Example trajectory in a gridworld where the agent can
        attempt to move right and down, although with a small probability
      the wind will move the agent down, independent of the action.\label{fig:gridworld}}
    \end{center}
  \end{subfigure}
  \hfill
  \begin{subfigure}[t]{0.45\linewidth}
    \begin{center}
      \scalebox{0.8}{ 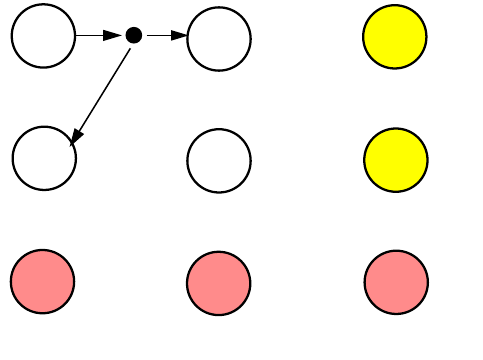 }
      \caption{
        PA describing the dynamics of Fig~\ref{fig:gridworld} as a
        1\nicefrac{1}{2} player game. The large circles indicate states (agent
        decisions) and the small black circles denote the
        environment response probabilities\label{fig:gridworld_mdp}.
      }
    \end{center}
  \end{subfigure}
  \caption{
    Example of gridworld probabilistic automata (PA).\label{fig:explicit_pa}
  }    

\end{figure}

Note that probabilistic automata are equivalently characterized as
\emph{1\nicefrac{1}{2} player games} where each round has the agent
choose an action and then the environment samples a state transition
outcome. In fact, this alternative characterization is implicitly
encoded in the directed bipartite graph used to visualize
probabilistic automata (see Fig~\ref{fig:gridworld_mdp}). In this
language, we refer to the nodes where the agent makes a decision as a
\textbf{decision node} and the nodes where the environment samples an
outcome as a \textbf{chance node}.

Next, we develop machinery to distinguish between desirable and
undesirable traces. For simplicity, we focus on finite trace
properties, referred to as specifications, that are decidable within
some fixed $\tau \in \Nat$ time steps, e.g., ``Recharge before t=20.''

\begin{mddef}
  A \textbf{task specification}, $\f$, (or simply
  \textbf{specification}) is a subset of traces. For simplicity, we
  shall assume that each trace is of a fixed length $\tau \in \Nat$,
  e.g.,
  \begin{equation}
    \varphi \subseteq (A \times S)^\tau 
  \end{equation}

  \noindent
  A collection of specifications, $\Phi$, is called a \textbf{concept
    class}.  Further, we define $\true \eqdef (A \times S)^\tau$,
  $\neg \varphi \eqdef \true \setminus \varphi$, and
  $\false \eqdef \neg \true$.
\end{mddef}

Often specifications are not directly given as sets, but induced by
abstract descriptions of a task. For example, the task ``avoid lava''
induces a concrete set of traces that never enter lava tiles. If the
workspace/world/dynamics change, this abstract specification would map
to a different set of traces.

\subsection{Specification Inference from Demonstrations:}
The primary task in this paper is to find the specification that best
explains/forecasts the behavior of an agent.  As in our prior
work~\cite{max_ent_spec_learning}, we formalize our problem statement
as:
\begin{mddef}\label{problem1}
  The \textbf{specification inference from demonstrations} problem is
  a tuple $(M, \Demos, \Phi, D)$ where $M=(S, s_0, A, \delta) $ is a
  probabilistic automaton, $\Demos$ is a (multi-)set of $\tau$-length
  traces drawn from an unknown distribution induced by a teacher
  attempting to demonstrate (satisfy) some unknown task specification
  within $M$, $ \Phi $ is a concept class of specifications, and $D$
  is a prior distribution over $\Phi$.  A solution to
  $(M, \Demos, \Phi, D)$ is:
  \begin{equation}
    \label{eq:RiskAverseProblem} \varphi^* \in \argmax_{\varphi \in \Phi}\Prob
    (\Demos~|~M, \varphi)\cdot \Pr_{\varphi \sim D}(\varphi)\\
  \end{equation}
  where $\Pr(\Demos~|~M, \varphi)$ denotes the likelihood that the
  teacher would have demonstrated $\Demos$ given the task $\varphi$.
\end{mddef}
Of course, by itself, the above formulation is ill-posed as
$\Pr(\Demos~|~M, \varphi)$ is left undefined. Below, we
shall propose leveraging Maximum Causal Entropy Inverse Reinforcement
Learning (IRL) to select the demonstration likelihood distribution in
a regret minimizing manner.

\section{Leveraging Inverse Reinforcement Learning}\label{sec:solution}

The key idea of Inverse Reinforcement Learning (IRL), or perhaps more
accurately Inverse Optimal Control, is to find the reward structure
that best explains the actions of a reward optimizing agent operating
in a Markov Decision Process. We formalize below.

\begin{mddef}
  \begin{flushleft}
    A \textbf{Markov Decision Process (MDP)} is a probabilistic
    automaton endowed with a \textbf{reward map} from states
    to reals, $r : S \to \Reals$. This reward mapping is
    lifted to traces via,
    \begin{equation}
      \label{eq:reward}
      R(\xi) \eqdef \sum_{s \in \xi}r(s).
    \end{equation}
  \end{flushleft}
\end{mddef}
\begin{remark}
  Note that a temporal discount factor, $\gamma \in [0, 1]$ can be
  added into~\eqref{reward} by introducing a sink state, $\$$, to the
  MDP, where $r(\$) = 0$ and
  \begin{equation}
    \label{eq:discount}
    \Pr(s' = \$~|~s, a) =
    \begin{cases}
      \gamma & \text{if } s \neq \$\\
      1 & \text{otherwise}
    \end{cases}.
  \end{equation}
\end{remark}
Given a MDP, the goal of an agent is to maximize the expected trace
reward.  In this work, we shall restrict ourselves to rewards that are
given as a linear combination of \textbf{state features}, $\mathbf{f}:
S \to \PosReals^n$, e.g.,
\begin{equation}
  r(s) = \mathbf{\theta} \cdot \mathbf{f}(s)
\end{equation}
for some $\mathbf{\theta} \in \Reals^n$. Note that
since state features can themselves be rewards, such a restriction
does not actually restrict the space of possible rewards.

\begin{example}
  Let the components of $\mathbf{f}(s)$ be distances to various
  locations on a map. Then the choice of $\mathbf{\theta}$
  characterizes the relative preferences in avoiding/reaching the
  respective locations.
\end{example}

Formally, we model an agent as acting according to a
\textbf{policy}.
\begin{mddef}
  \begin{flushleft}
    A policy, $\pi$, is a state indexed distribution over actions,
    \begin{equation}
      \Pr(a~|~s) = \pi(a~|~s).
    \end{equation}
  \end{flushleft}
\end{mddef}
In this language, the agent's goal is equivalent to finding a policy
which maximizes the expected trace reward. We shall refer to a trace
generated by such an agent as a \textbf{demonstration}. Due to the
Markov requirement, the likelihood of a demonstration, $\xi$, given a particular
policy, $\pi$, and probabilistic automaton, $M$, is easily stated as:
\begin{equation}
  \Pr(\xi~|~M, \pi) = \prod_{s', a, s \in \xi}\Pr(s'~|~s, a)\cdot \Pr(a~|~s).
\end{equation}
Thus, the likelihood of multi-set of i.i.d demonstrations, $X$,
is given by:
\begin{equation}\label{eq:demo_likelihoods}
  \Pr(X~|~M, \pi) = \prod_{\xi \in X}\Pr(\xi~|~M, \pi).
\end{equation}

\subsection{Inverse Reinforcement Learning (IRL):}
As previously stated, the main motivation in introducing the MDP
formalism has been to discuss the inverse problem.  Namely, given a
set of demonstrations, find the reward that best ``explains'' the
agent's behavior, where by ``explain'' one typically means that under
the conjectured reward, the agent's behavior was approximately
optimal. Notice however, that many undesirable rewards satisfy this
property. For example, consider the following reward in which every
demonstration is optimal,
\begin{equation}\label{eq:r_0}
  r: s \mapsto 0.
\end{equation}
Furthermore, observe that given a fixed reward, many policies are
approximately optimal! For instance, using~\eqref{r_0}, an optimal
agent could pick actions uniformly at random or select a single
action to always apply.

\subsection{Maximum Causal Entropy IRL:}

A popular, and in practice effective, solution to the lack of unique
policy conundrum is to appeal to the \textbf{principle of maximum
causal entropy}~\cite{mce}. To formalize this principle, we recall the
definitions of causally conditioned
probability~\cite{kramer1998directed} and causal
entropy~\cite{kramer1998directed}\cite{permuter2008directed}.
\begin{mddef}
  Let $X_{1:\tau} \eqdef X_{1}, \ldots, X_\tau$ denote a temporal sequence
  of $\tau \in \Nat$ random variables. The probability of a sequence $Y_{1:\tau}$
  \textbf{causally conditioned} on sequence $X_{1:\tau}$ is:
  \vspace{-1.2em}
  \begin{equation}
    \Pr(Y_{1:\tau}~||~X_{1:\tau}) \eqdef \prod_{t=1}^\tau \Pr(Y_t~|~X_{1:t}, Y_{1:{t-1}})
  \end{equation}
  \vspace{-1.2em}
  
  \noindent
  The \textbf{causal entropy} of $Y_{1:\tau}$ given $X_{1:\tau}$ is defined as,
  \begin{equation}
    H(Y_{1:\tau}~||~X_{1:\tau}) \eqdef \E_{Y_{1:\tau}, X_{1:\tau}}[-\log(\Pr(Y_{1:\tau}~||~X_{1:\tau}))]
  \end{equation}
\end{mddef}

In the case of inverse reinforcement learning, the principle
of maximum causal entropy suggests forecasting using the policy whose
action sequence, $A_{1:\tau}$, has the highest causal entropy, conditioned
on the state sequence, $S_{1:\tau}$. That is, find the policy that maximizes
\begin{equation}
  H(A_{1:\tau}~||~S_{1:\tau}),
\end{equation}
subject to feature matching constraints, $\E[\mathbf{f}]$, e.g., does the resulting
policy, $\pi^*$, complete the task as seen in the data. 
Compared to all other policies, this policy (i) minimizes regret with
respect to model/reward uncertainty, (ii) ensures that the agent's
predicted policy does not depend on the future, (iii) is consistent
with observed feature statistics~\cite{mce}.

Concretely, as proved in~\cite{mce}, when an agent is attempting to
maximize the sum of feature state rewards, $\sum_{t=1}^T\mathbf{\theta}
\cdot \mathbf{f}(s_t)$, the principle of maximum causal entropy
prescribes the following policy:

\begin{mdframed}
  \textbf{Maximum Causal Entropy Policy:}
  \begin{equation}
    \log\big(\pi_{\mathbf{\theta}}(a_t~|~s_t)\big) \eqdef Q_{\mathbf{\theta}}(a_t, s_t) - V_{\mathbf{\theta}}(s_t)
  \end{equation}
  \vspace{-2em}
  where
  \begin{equation}\label{eq:soft_bellman_backup}
    \begin{split}
      &Q_{\mathbf{\theta}}(a_t, s_t) \eqdef \E_{s_{t+1}}\left[ V_{\mathbf{\theta}}(s_{t+1})~|~s_t, a_t\right] + \mathbf{\theta} \cdot \mathbf{f}(s_t) \\
      &V_{\mathbf{\theta}}(s_t) \eqdef \ln \sum_{a_t} e^{Q_{\mathbf{\theta}}(a_t, s_t)} \eqdef \softmax_{a_t} Q_{\mathbf{\theta}}(a_t, s_t).
  \end{split}
\end{equation}
  
\end{mdframed}
where, $\theta$ is such that~\eqref{soft_bellman_backup} results in a policy
which matches feature demonstrations.
\begin{remark}
  Note that replacing softmax with max in~\eqref{soft_bellman_backup}
  yields the standard Bellman Backup~\cite{bellman1957dynamic} used to
  compute the optimal policy in tabular reinforcement learning. Further, it
  can be shown that maximizing causal entropy corresponds to believing that the
  agent is exponentially biased towards high reward policies~\cite{mce}:
  \begin{equation}\label{eq:policy_dist}
    \Pr(\pi_\theta~|~M) \propto \exp\Big(\E_{\substack{\xi}}[R_\theta(\xi)~|~\pi_\theta, M]\Big),
  \end{equation}
  where~\eqref{soft_bellman_backup} is the most likely policy
  under~\eqref{policy_dist}.
\end{remark}

\begin{remark}
  In the special case of scalar state features,
  $\mathbf{f}: S \to \PosReals$, the maximum causal entropy
  policy~\eqref{soft_bellman_backup} becomes increasingly optimal as
  $\theta \in \Reals$ increases (since softmax monotonically
  approaches max). In this setting, we shall refer to $\theta$ as the
  agent's \textbf{rationality coefficient}.
\end{remark}

\subsection{Non-Markovian Rewards:}
The MDP formalism traditionally requires that the reward map be
Markovian (i.e., state based); however, in practice, many
tasks are history dependent, e.g. touch a red tile and
then a blue tile.

A common trick within the reinforcement learning literature is to
simply change the MDP and add the necessary history to the state so
that the reward is Markovian, e.g. a flag for touching a red tile.
However, in the case of inverse reinforcement learning, by definition,
one does not know what the reward is. Therefore, one cannot assume to
a priori know what history suffices.

Further exacerbating the situation is the fact that na\"ively
including the entire history into the state results in an exponential
increase in the number of states. Nevertheless, as we shall soon see, by
restricting the class of rewards to represent task specifications,
this curse can be mitigated to only result in a blow-up that is at
most \textbf{linear} in the state space size and in the trace length!

To this end, we shall find it fruitful to develop machinery for
embedding the full trace history into the state space.  Explicitly, we
shall refer to the process of adding all history to a probabilistic
automaton's (or MDP's) state as \textbf{unrolling}.
\begin{mddef}\label{def:unroll}
  Let $M = (S, s_0, A, \delta)$ be a PA.  The \textbf{unrolling} of
  $M$ is a PA, $M'= (S', s_0, A, \delta')$, where
  \vspace{-0.3em}
    \begin{equation}
      \begin{split}
        S' = \{s_0\} \times \bigcup_{i=0}^\infty (A \times S)^i
        \hspace{5em}
        \delta'(\xi_{n+1}, a, \xi_{n}) = \delta(s_{n+1}, a, s_n)\\
        \xi_n = \Big(s_0, \ldots, (a_{n-1}, s_{n})\Big)\hspace{5.5em}
        \xi_{n+1} = \Big(s_0,\ldots,  (a_n, s_{n+1})\Big)
    \end{split}
  \end{equation}
\end{mddef}    

If $R : S^\tau \to \Reals$ is a non-Markovian reward over
$\tau$-length traces, then we endow the corresponding unrolled
PA with the now Markovian Reward,

\begin{equation}\label{eq:unrolled_reward} r'\Big(s_0, \ldots, (a_{n-1}, s_n)\Big)
  \eqdef
  \begin{cases} R(s_0, \ldots, s_n) & \text{if } n = \tau\\ 0 &
    \text{otherwise}
  \end{cases}.
\end{equation}
Further, by construction the reward is Markovian in $S'$ and only
depends only $\tau$-length state sequences,
\begin{equation}
  \sum_{t=0}^\infty r'\big((s_0, a_0), \ldots s_\tau\big) = R(s_0, \ldots, s_\tau).
\end{equation}

\begin{figure}[h]
  \begin{center}
    \scalebox{0.8}{ 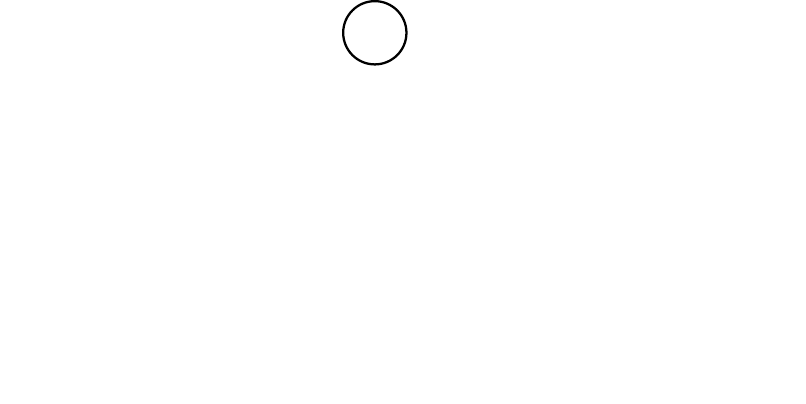 }
    \caption{Decision tree generated by the PA shown in~\figref{explicit_pa} and specification ``By $\tau=2$, reach a yellow
      tile while avoiding red tiles.''. Here a binary reward is given
      depending on whether or not the agent satisfies the specification\label{fig:dtree}.}
  \end{center}
\end{figure}

Next, observe that for $\tau$-length traces, the 1\nicefrac{1}{2}
player game formulation's bipartite graph forms a tree of depth
$\tau$ (see Fig~\ref{fig:dtree}). Further, observe that each leaf corresponds to unique $\tau$-length
trace. Thus, to each leaf, we associate the corresponding trace's reward,
$R(\xi)$. We shall refer to this tree as a \textbf{decision tree},
denoted $\mathbb{T}$.

\vspace{-2em}
\vspace{-2.5em}

Finally, observe that the trace reward depends only on the sequence of
agent actions, $A$, and environment actions, $A_e$. That is, $\mathbb{T}$ can be interpreted as a
function:
\begin{equation}
  \mathbb{T} : {(A \times A_e)^\tau} \to \Reals.
\end{equation}

\subsection{Specifications as Non-Markovian Rewards:}
Next, with the intent to frame our specification inference problem as
an inverse reinforcement learning problem, we shall overload notation
and denote by $\varphi$ the following non-Markovian reward
corresponding to a specification $\varphi \in (A \times S)^\tau$,

\begin{equation}
  \varphi(\xi) \eqdef
  \begin{cases}
      1 & \text{if } \xi \in \varphi\\
      0 & \text{otherwise}
  \end{cases}
  .
\end{equation}
Note that the corresponding decision tree is then a Boolean predicate:
\begin{equation}
  \mathbb{T}_\varphi : {(A \times A_e)^\tau} \to \{0, 1\}.
\end{equation}

\subsection{Computing Maximum Causal Entropy Specification Policies:}

Now let us return to the problem of computing the policy prescribed
by~\eqref{soft_bellman_backup}. 
In particular, note that viewing
the unrolled reward~\eqref{unrolled_reward} as a scalar state
feature results in the following soft-Bellman Backup:
\begin{equation}\label{eq:soft_bellman_spec}
  \begin{split}
    &Q_{\mathbf{\theta}}(a_t, \xi_t) = \E\left[ V_{\mathbf{\theta}}(s_{t+1})~|~\xi_t, a_t\right]\\
    &V_{\mathbf{\theta}}(\xi_t) =
    \begin{cases}
      \theta\cdot \varphi(\xi_t) & \text{if } t = \tau\\
      \softmax_{a_t} Q_{\mathbf{\theta}}(a_t, \xi_t) & \text{otherwise}
    \end{cases}
  \end{split},
\end{equation}
where $\xi_i \in \{s_0\}\times (A\times S)^i$ denotes a state in the unrolled MDP.

\vspace{1em}
Eq~\eqref{soft_bellman_spec} thus suggests a
na\"ive dynamic programming scheme over $\mathbb{T}$ starting at the
$t=\tau$ leaves to compute $Q_\theta$ and $V_\theta$ (and thus
$\pi_{\mathbf{\theta}}$).

\begin{wrapfigure}[13]{r}{6.4cm}
  \vspace{-3em}
  \begin{center}
    \scalebox{0.8}{ 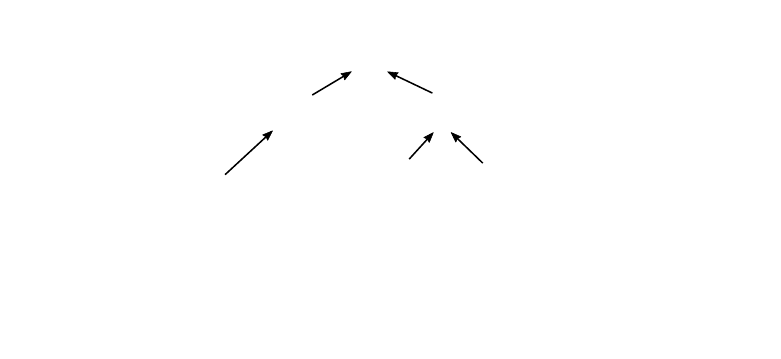 }
    \caption{Computation graph generated from
      applying~\eqref{soft_bellman_backup} to
      the decision tree shown in Fig~\ref{fig:dtree}. Here smax and avg denote the softmax and weighted average respectively.\label{fig:cgraph_1}}
  \end{center}
\end{wrapfigure}

Namely, in $\mathbb{T}$, the chance nodes, which correspond to action/state
pairs, are responsible for computing $Q$ values and the decision
nodes, which correspond to states waiting for an action to be applied,
are responsible for computing $V$ values. For chance nodes this is
done by taking the $\softmax$ of the values of the child
nodes. Similarly, for decision nodes, this is done by taking a
weighted average of the child nodes, where the weights correspond to
the probability of a given transition. This, at least conceptually,
corresponds to transforming $\mathbb{T}$ into a bipartite computation graph
(see Fig~\ref{fig:cgraph_1}).

Next, note that (i) the above dynamic programming scheme can be
trivially modified to compute the expected trace reward of
the maximum causal entropy policy and (ii) the expected reward increases
\footnote{Formally, this is due to (a) softmax and average being monotonic (b) trajectory rewards only increasing with 
$\theta$, and (c) $\pi$ exponentially biasing towards high Q-values.}
with the rationality coefficient $\theta$.

Observe then that, due to monotonicity, bisection (binary search)
approximates $\theta$ to tolerance $\epsilon$ in $O(\log(1/\epsilon))$
time. Additionally, notice that the likelihood of each demonstration
can be computed by traversing the path of length $\tau$ in
$\mathbb{T}$ corresponding to the trace and multiplying the
corresponding policy and transition
probabilities~\eqref{demo_likelihoods}.  Therefore, if $|A_e| \in
\Nat$ denotes the maximum number of outcomes the environment can
choose from (i.e, the branching factor for chance nodes), it follows
that the run-time of this na\"ive scheme is:
\begin{equation}\label{eq:runtime_naive}
  O\bigg(\overbrace{\underbrace{\Big(|A|\cdot |A_e|\Big)^\tau}_{|\mathbb{T}|} ~\cdot \underbrace{\log(1/\epsilon)}_{\text{Feature Matching}}}^{\text{compute policy}} + \underbrace{\tau|\Demos|}_{\text{evaluate demos}}\bigg).
\end{equation}

\subsection{Task Specification Rewards:}

Of course, the problem with this na\"ive approach is that explicitly
encoding the unrolled tree, $\mathbb{T}$, results in an exponential
blow-up in the space and time complexity. The key insight in this
paper is that the additional structure of task specifications enables
avoiding such costs while still being expressive. In particular, as
is exemplified in Fig~\ref{fig:cgraph_1}, the computation graphs
for task specifications are often highly redundant and apt for
compression.

\begin{wrapfigure}[7]{r}{4cm}
  \vspace{-1em}
  \centering
  \scalebox{0.8}{ 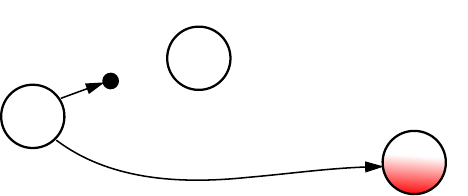}
  \caption{Reduction of the decision
    tree shown in~\figref{dtree}.
    \label{fig:rodd}}
\end{wrapfigure}


In particular, we shall apply the following two semantic preserving
transformations: (i) Eliminate nodes whose children are isomorphic
sub-graphs, i.e., inconsequential decisions (ii) Combine all
isomorphic sub-graphs i.e., equivalent decisions. We refer to the
limit of applying these two operations as a \textbf{reduced ordered
probabilistic decision diagram} and shall denote\footnote{Mnemonic: $\mathcal{T}$ is
a (typographically) slimmed down variant of $\mathbb{T}$} the reduced
variant of $\mathbb{T}$ as $\mathcal{T}$.

\begin{remark}
  For those familiar, we emphasize that these decision diagrams are
  MDPs, not Binary Decision Diagrams (see
  \secref{robdd}). Importantly, more than two actions can be taken
  from a node if $\max(|A|, |A_e|) \geq 2$ and $A_e$ has a state
  dependent probability distribution attached to it.  That said, the
  above transformations are \textbf{exactly} the reduction rules for
  BDDs~\cite{bryant1992symbolic}.
\end{remark}

As Fig~\ref{fig:rodd} illustrates, reduced decision diagrams can be
much smaller than their corresponding decision tree. Nevertheless,
we shall briefly postpone characterizing $|\mathcal{T}|$ until
developing some additional machinery in~\secref{robdd}.
Computationally, three problems remain.
\begin{enumerate}
\item How can our na\"ive dynamic programming scheme be adapted to
  this compressed structure. In particular, because many interior nodes
  have been eliminated, one must take care when
  applying~\eqref{soft_bellman_spec}.
\item How do concrete demonstrations map to paths in the compressed
  structure when evaluating likelihoods~\eqref{demo_likelihoods}.
\item How can one construct $\mathcal{T}$ without first constructing
  $\mathbb{T}$, since failing to do so would negate any complexity savings.
\end{enumerate}
We shall postpone discussing solutions to the second and third
problems until~\secref{robdd}.  The first problem however, can readily
be addressed with the tools at hand. Recall that in the variable
ordering, nodes alternate between decision and chance nodes (i.e.,
agent and environment decisions), and thus alternate between taking a
softmax and expectations of child values
in~\eqref{soft_bellman_spec}. Next, by definition, if a node is
skipped in $\mathcal{T}$, then it must have been inconsequential. Thus
the trace reward must have been independent of the decision made at
that node. Therefore, the softmax/expectation's
corresponding to eliminated nodes must have been over a constant value
- otherwise the eliminated sequences would be distinguishable w.r.t
$\varphi$.  The result is summarized in the following identities,
where $\alpha$ denotes the value of an eliminated node's children.

\vspace{-0.5em}
\begin{equation}\label{eq:eliminated_value}
    \softmax(\overbrace{\alpha, \ldots, \alpha}^{|A|}) = \log(e^\alpha + \ldots + e^\alpha ) = \ln(|A|) + \alpha
\end{equation}
\begin{equation}
    \label{eq:2}
    \E_x[\alpha] = \sum_{x}p(x)\alpha = \alpha
\end{equation}

Of course, it could also be the case that a sequence of nodes is
skipped in $\mathcal{T}$. Using~\eqref{eliminated_value}, one can
compute the change in value, $\Delta$, that the eliminated sequence of
$n$ decision nodes and any number of chance nodes would have applied
in $\mathbb{T}$:
\begin{equation}\label{eq:eliminated_value_seq}
  \Delta(n, \alpha) = \ln(|A|^n) + \alpha = n \ln(|A|) + \alpha
\end{equation}
Crucially, evaluation of this compressed computation graph is linear
in $|\mathcal{T}|$ which as shall later prove, is often much smaller
than $|\mathbb{T}|$.

\section{Constructing and Characterizing $\mathcal{T}$}\label{sec:robdd}

Let us now consider how to avoid the construction of $\mathbb{T}$ and
characterize the size of the reduced ordered decision diagram,
$\mathcal{T}$.  We begin by assuming that the underlying dynamics is
well-approximated in the random-bit model.

\begin{mddef}
  For $q \in \Nat$, let $\mathbf{c} \sim \boolcube{q}$ denote
  the random variable representing the result of flipping $q \in \Nat$
  fair coins.
  We say a probabilistic automata $ M = (S, s_0, A, \delta) $ is
  $(\epsilon, q)$ approximated in the random bit model if there exists a
  mapping,
  \begin{equation}
    \hat{\delta} : S \times A \times \boolcube{q} \to S
  \end{equation}
  such that for all $s, a, s' \in S\times A\times S$:
  \begin{equation}\label{eq:approx_delta}
    \left|~\delta(s, a, s') - \Pr_{\mathbf{c}\sim \boolcube{q}}\left (\hat{\delta}(s, a, c) = s'\right )\right| \leq \epsilon.
  \end{equation}
\end{mddef}

For example, in our gridworld example (Fig~\ref{fig:gridworld}), if
$\mathbf{c} \in \setof{0, 1}^3$, elements of $s$ are interpreted as pairs
in $\Reals^2$, and the right/down actions are interpreted as the
addition of the unit vectors $(1, 0)$ and $(0, 1)$ then,
\begin{equation}\label{eq:random_bit_example}
  \hat{\delta}(s, a, \mathbf{c}) =
  \begin{cases}
    s & \text{if } \max_i[(s + a)_i] > 1\\
    s + (0, 1) & \text{else if  } \mathbf{c} = 0\\
    s + a & \text{otherwise}
  \end{cases},
\end{equation}
As can be easily confirmed,~\eqref{random_bit_example}
satisfies~\eqref{approx_delta} with $\epsilon = 0$. In the sequel, we
shall take access to $\hat{\delta}$ as given\footnotemark. Further, to
simplify exposition, until Sec~\ref{sec:valid}, we shall additionally
require that the number of actions, $|A|$, be a power of $2$. This
assumption implies that $A$ can be encoded using exactly
$\log_2(|A|)$ bits.

\footnotetext{See~\cite{model_counters_guide} for an explanation on systematically
deriving such encodings.
}

Under the above two assumptions, the key observation is to recognize
that $\mathbb{T}$ (and thus $\mathcal{T}$) can be viewed as a Boolean
predicate over an alternating sequence of action bit strings and coin
flip outcomes determining if the task specification is satisfied, i.e.,
\begin{equation}
    \mathbb{T} : \{0, 1\}^{n} \to \{0, 1\},
\end{equation}
where $n \eqdef \tau\cdot \log_2(|A \times A_e|) = \tau\cdot (q + \log_2(|A|))$.
That is to say, the resulting decision diagram can be re-encoded as a
reduced ordered \textbf{binary} decision diagram~\cite{bryant1992symbolic}.

\begin{mddef}
  A \textbf{reduced ordered binary decision diagram} (BDD), is a
  representation of a Boolean predicate $h(x_1, x_2, \ldots, x_n)$ as
  a reduced ordered (deterministic) decision diagram, where each
  decision corresponds to testing a bit $x_i \in \setof{0, 1}$.  We
  denote the BDD encoding of $\mathcal{T}$ as $\mathcal{B}$.
\end{mddef}

Binary decision diagrams are well developed both in a
theoretical and practical sense. 
Before exploring these
benefits, we first note that this change has introduced an additional
problem. First, note that in $\mathcal{B}$, decision and
chance nodes from $\mathbb{T}$ are now encoded as sequences of decision and
chance nodes. For example, if $a \in A$ is encoded by the 4-length bit
sequence $b_1b_2b_3b_4$, then four decisions are made by the agent
before selecting an action. Notice however that the original semantics
are preserved due to associativity of the $\softmax$ and $\E$
operators. In particular, recall that by definition,
\begin{equation}
  \begin{split}
    \softmax(&\alpha_1, \ldots, \alpha_4) = \ln(\sum_{i=1}^4 e^{\alpha_i})
    = \ln(e^{\ln(e^{\alpha_1}+ e^{\alpha_2})} + e^{\ln(e^{\alpha_3}+ e^{\alpha_4})})\\
    &\eqdef \softmax(\softmax(\alpha_1, \alpha_2), \softmax(\alpha_3, \alpha_4))
  \end{split}
\end{equation}
and thus the semantics of the sequence decision nodes is equivalent to
the decision node in $\mathbb{T}$. Similarly, recall that the coin flips are
fair, and thus expectations are computed via $\text{avg}(\alpha_1,
\ldots, \alpha_n) = \nicefrac{1}{n}(\sum_{i=1}^n
\alpha_i)$. Therefore, averaging over two sequential coin flips
yields,
\begin{equation}
  \begin{split}
    \text{avg}(\alpha_1, \ldots, \alpha_4) &\eqdef \frac{1}{4}\sum_{i=1}^4 \alpha_i = \frac{1}{2}(\frac{1}{2}(\alpha_1 + \alpha_2) + \frac{1}{2}(\alpha_3 + \alpha_4))\\
    &\eqdef \text{avg}(\text{avg}(\alpha_1, \alpha_2), \text{avg}(\alpha_3, \alpha_4))
  \end{split}
\end{equation}
which by assumption~\eqref{approx_delta}, is the same as applying
$\E$ on the original chance node. Finally, note that skipping over
decisions needs to be adjusted slightly to account for sequences of
decisions. Recall that via~\eqref{eliminated_value_seq},
the corresponding change in value, $\Delta$, is a function of initial
value, $\alpha$, and the number of agent actions skipped, i.e., $|A|^n$ for
$n$ skipped decision nodes. Thus, in the BDD, since each decision
node has two actions, skipping $k$ decision bits corresponds to skipping
$2^k$ actions. Thus, if $k$ decision bits are skipped over in the BDD,
the change in value, $\Delta$, becomes,
\begin{equation}
  \Delta(k, \alpha) = \alpha + k\ln(2).
\end{equation}
Further, note that $\Delta$ can be computed in constant time while
traversing the BDD. Thus, the dynamic programming scheme is linear
in the size of $\mathcal{B}$.

\subsection{Size of $\mathcal{B}$:}

\begin{figure}[h]
  \centering
  \scalebox{0.7}{ 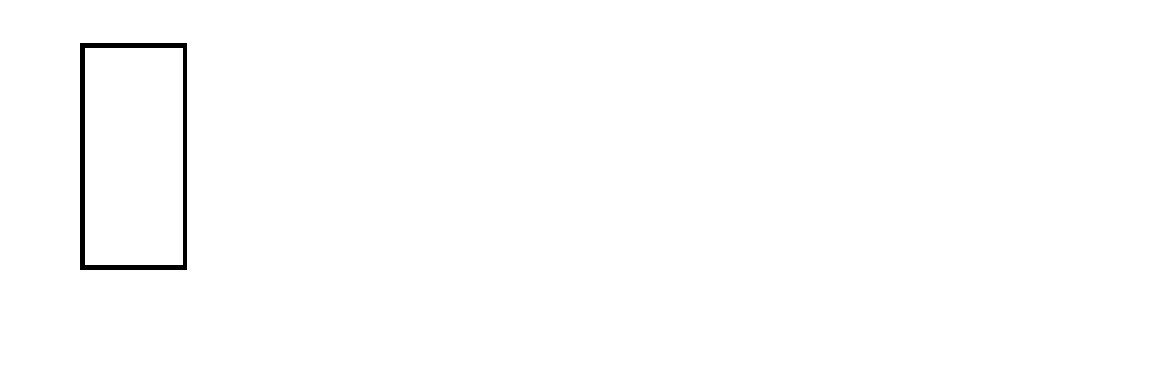}
  \caption{Generic network of Boolean modules for which
    \thmref{bdd_size} holds.}
  \label{fig:linear_model_of_computation}
\end{figure}

Next we return to the question of how big the compressed decision
diagram can actually be. To this aim, we cite the following
(conservative) bound on the size of an BDD given an encoding of the
corresponding Boolean predicate in the linear model computation
illustrated in Fig~\ref{fig:linear_model_of_computation} (for more
details, we refer the reader to \cite{knuth2009art}).

\vspace{-.8em}
\vspace{-1.5em}

\noindent
In particular, consider an arbitrary Boolean predicate
\begin{equation}
  f: \boolcube{n} \to \boolcube{}
\end{equation}
and a sequential arrangement of $n$ Boolean modules,
$f_1, f_2, \ldots, f_n$ where each $f_i$ has shape: 
\begin{equation}
  f_i : \boolcube{} \times \boolcube{a_{i-1}} \times \boolcube{b_i} \to \boolcube{a_i} \times \boolcube{b_{i-1}},
\end{equation}
and takes as input $x_i$ as well as $a_{i-1}$ outputs of its left
neighbor and $b_i$ outputs of the right neighbor ($b_0 = 0, a_n = 1$).
Further, assume that this arrangement is well defined, e.g. for each
assignment to $x_1, \ldots, x_n$ there exists a unique way to set each
of the inter-module wires. We say these modules compute $f$ if the
final output is equal to $f(x_1, \ldots, x_n)$. 
\begin{mdtheo}\label{thm:bdd_size}
  If $f$ can be computed by a linear arrangement of such modules,
  ordered $x_1, x_2, \ldots, x_n$, then the
  size, $S \in \Nat$, of its BDD (in the same order), is upper bounded~\cite{bryant1992symbolic} by:
  \vspace{-0.5em}
  \begin{equation}\label{eq:bdd_bound}
    S \leq \sum_{k=1}^n 2^{a_k \cdot \left(2^{b_k} \right)}.
  \end{equation}
\end{mdtheo}

To apply this bound to our problem, recall that $\mathcal{B}$ computes
a Boolean function where the decisions are temporally ordered and
alternate between sequences of agent and environment decisions.  Next,
observe that because the traces are bounded (and all finite sets are
regular), there exists a finite state machine which can monitor the
satisfaction of the specification.

\begin{remark}
  In the worst case, the monitor could be the unrolled decision tree,
  $\mathbb{T}$.  This monitor would have exponential number of
  states. In practice, the composition of the dynamics
  and the monitor is expected to be much smaller.
\end{remark}

\begin{wrapfigure}[10]{r}{5.5cm}
  \vspace{-2em}
  \centering
  \scalebox{0.7}{ 
\begingroup%
  \makeatletter%
  \providecommand\color[2][]{%
    \errmessage{(Inkscape) Color is used for the text in Inkscape, but the package 'color.sty' is not loaded}%
    \renewcommand\color[2][]{}%
  }%
  \providecommand\transparent[1]{%
    \errmessage{(Inkscape) Transparency is used (non-zero) for the text in Inkscape, but the package 'transparent.sty' is not loaded}%
    \renewcommand\transparent[1]{}%
  }%
  \providecommand\rotatebox[2]{#2}%
  \newcommand*\fsize{\dimexpr\f@size pt\relax}%
  \newcommand*\lineheight[1]{\fontsize{\fsize}{#1\fsize}\selectfont}%
  \ifx\svgwidth\undefined%
    \setlength{\unitlength}{204.11677551bp}%
    \ifx\svgscale\undefined%
      \relax%
    \else%
      \setlength{\unitlength}{\unitlength * \real{\svgscale}}%
    \fi%
  \else%
    \setlength{\unitlength}{\svgwidth}%
  \fi%
  \global\let\svgwidth\undefined%
  \global\let\svgscale\undefined%
  \makeatother%
  \begin{picture}(1,0.4888616)%
    \lineheight{1}%
    \setlength\tabcolsep{0pt}%
    \put(0,0){\includegraphics[width=\unitlength,page=1]{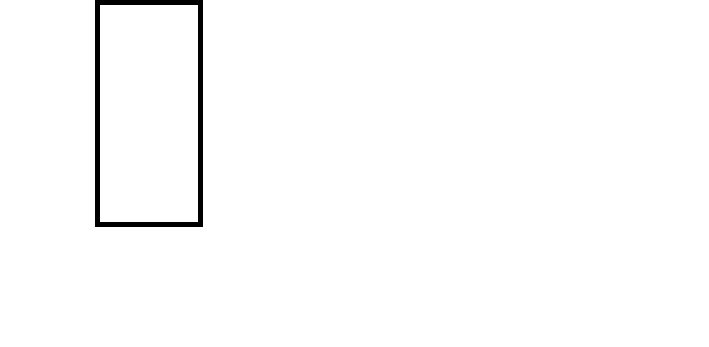}}%
    \put(0.1878726,0.31810107){\color[rgb]{0,0,0}\makebox(0,0)[lt]{\lineheight{1.25}\smash{\begin{tabular}[t]{l}$f_2$\end{tabular}}}}%
    \put(0,0){\includegraphics[width=\unitlength,page=2]{linear_model2.pdf}}%
    \put(0.4637293,0.33748038){\color[rgb]{0,0,0}\makebox(0,0)[lt]{\lineheight{1.25}\smash{\begin{tabular}[t]{l}$\log_2(2^q |A \times S \times S_\varphi|)$\end{tabular}}}}%
    \put(0.05235257,0.00746021){\color[rgb]{0,0,0}\makebox(0,0)[lt]{\lineheight{1.25}\smash{\begin{tabular}[t]{l}action or coin flip\end{tabular}}}}%
    \put(0,0){\includegraphics[width=\unitlength,page=3]{linear_model2.pdf}}%
    \put(0.61138131,0.16493322){\color[rgb]{0,0,0}\makebox(0,0)[lt]{\lineheight{1.25}\smash{\begin{tabular}[t]{l}Current State + \\Partial Action + \\Partial Coin Flip\end{tabular}}}}%
    \put(0,0){\includegraphics[width=\unitlength,page=4]{linear_model2.pdf}}%
  \end{picture}%
\endgroup%
}
  \caption{Generic module in linear model of computation for
    $\mathcal{B}$.  Note that backward edges are not required.
  \label{fig:corr_picture}}
\end{wrapfigure}

Further, note that because this composed system is causal, no backward
wires are needed, e.g., $\forall k~.~b_k = 0$. In particular, observe
that because the composition of the dynamics and the monitor is
Markovian, the entire system can be uniquely described using the
monitor/dynamics state and agent/environment action
(see~\figref{corr_picture}).  This description can be encoded in
$\log_2(2^q|A \times S \times S_\varphi|)$ bits, where $q$
denotes the number of coin flips tossed by the environment and
$S_\varphi$ denotes the monitor state. Therefore, $a_k$ is upper
bounded by $\log_2(2^q |A \times S \times S_\varphi|)$. Combined
with~\eqref{bdd_bound} this results in the following bound on the size
of $\mathcal{B}$.

\begin{mdcorr}\label{corr:bdd_size}
  Let $M = (S, s_0, A, \delta)$ be a probabilistic automaton whose
  probabilistic transitions can be approximated using $q$ coin flips
  and let $\varphi$ be a specification defined for horizon $\tau$
  and monitored by a finite automaton with states $S_\varphi$.
  The corresponding BDD, $\mathcal{B}$, has size bounded by:
  \begin{equation}\label{eq:bound2}
    |\mathcal{B}| \leq \overbrace{\tau \cdot  \big( \log(|A|) + q \big)}^{\text{\# inputs}} \cdot \overbrace{\big( 2^q| A \times S \times S_\varphi| \big)}^{\text{bound on } 2^{a_k}}
  \end{equation}
\end{mdcorr}

Notice that the above argument implies that as the episode length
grows, $|\mathcal{B}|$ grows linearly in the horizon/states and
quasi-linearly in the agent/environment actions!

\begin{remark}
  Note that this bound actually holds for the minimal representation
  of the composed dynamics/monitor (even if it's unknown
  a-prori!). For example, if the property is $\true$, the BDD
  requires only one state (always evaluate true).  This also
  illustrates that the above bound is often very conservative. In particular, note that
  for $\varphi = \true$, $|\mathcal{B}| = 1$ , independent of the
  horizon or dynamics. However, the above bound will always be linear
  in $\tau$. In general, the size of the BDD will depend on the
  particular symmetries compressed.
\end{remark}

\begin{remark}
  With hindsight, corollary~\ref{corr:bdd_size} is not too
  surprising. In particular, if the monitor is known,
  then one could explicitly compose the dynamics MDP
  with the monitor, with the resulting MDP having at most
  $|S \times S_\varphi|$ states. If one then includes
  the time step in the state, one could perform the
  soft-Bellman Backup directly on this automaton.
  In this composed automaton each (action, state)
  pair would need to be recorded. Thus, one
  would expect $O(|S \times S_\varphi\times A|)$ space
  to be used.
  In practice, this explicit representation is much bigger than $\mathcal{B}$
  due to the BDDs ability to skip over time steps and
  automatically compress symmetries.
\end{remark}

\subsection{Constructing $\mathcal{B}$:}\label{sec:construct}
One of the biggest benefits of the BDD representation of a Boolean
function is the ability to build BDDs from a Boolean combinations of
other BDDs. Namely, given two BDDs with $n$ and $m$ nodes
respectively, it is well known that the conjunction or disjunction of
the BDDs has at most $n\cdot m$ nodes. Thus, in practice, if the
combined BDD's remain relatively small, Boolean combinations remain
efficient to compute and one does not construct the full
binary decision tree! Further, note that BDDs support function
composition. Namely, given predicates $f(x_1, \ldots, x_n)$ and
$n$ predicates $g_i(y_1, \ldots, y_k)$ the function
\begin{equation}
  f\bigg(g_1(y_1, \ldots, y_k), \ldots, g_n(y_1, \ldots, y_k)\bigg)
\end{equation}
can be computed in time~\cite{knuth2009art}:
\begin{equation}
  O(n\cdot |B_f|^2\cdot \max_i{|B_{g_i}|}),
\end{equation}
 where $B_f$ is the BDD for $f$ and $B_{g_i}$ are the BDDs for $g_i$.
Now, suppose $\hat{\delta}_1, \ldots \hat{\delta}_{\log(|S|)}$ are
Boolean predicates such that:
\begin{equation}
  \hat{\delta}(\mathbf{s}, \mathbf{a}, \mathbf{c}) = (\hat{\delta}_1(\mathbf{s}, \mathbf{a}, \mathbf{c}), \ldots, \hat{\delta}_{\log(|S|)}(\mathbf{s}, \mathbf{a}, \mathbf{c})).
\end{equation}
Thm~\ref{thm:bdd_size} and an argument similar to that for
Corr~\ref{corr:bdd_size} imply then that constructing $\mathcal{B}$,
using repeated composition, takes time bounded by a low degree
polynomial in $|A \times S\times S_\varphi|$ and the
horizon. Moreover, the space complexity before and after composition
are bounded by Corr~\ref{corr:bdd_size}.

\subsection{Evaluating Demonstrations:}\label{sec:demo}

Next let us return to the question of how to evaluate the likelihood
of a concrete demonstration in our compressed BDD. The key problem
is that the BDD can only evaluate (binary) sequences of actions/coin
flips, where as demonstrations are given as sequences of action/state
pairs.  That is, we need to algorithmically perform the following
transformation.
\begin{equation}
  s_0\mathbf{a}_0\mathbf{s}_1\ldots \mathbf{a}_{n}\mathbf{s}_{n+1} \mapsto \mathbf{a}_1\mathbf{c}_1\ldots \mathbf{a}_n \mathbf{c}_n
\end{equation}
Given the random bit model assumption, this transformation can be
rewritten as a series of Boolean Satisifability problems:
\begin{equation}\label{eq:find_coins}
  \exists~ \mathbf{c}_i~.~\hat{\delta}(\mathbf{s}_i, \mathbf{a}_i, \mathbf{c}_i) = \mathbf{s}_{i+1}
\end{equation}
While potentially intimidating, in practice such problems are quite
simple for modern SAT solvers, particularly if the number of coin
flips used is small. Furthermore, many systems are translation
invariant. In such systems, the results of a single query~\eqref{find_coins},
can be reused on other queries. For example,
in~\eqref{random_bit_example}, $\mathbf{c} = \mathbf{0}$ always results in
the agent moving to the right.  Nevertheless, in general, if $q$ coin
flips are used, encoding all the demonstrations takes at most
$O(|X|\cdot\tau\cdot 2^q)$, in the worst case.

\subsection{Run-time analysis:}
We are finally ready to provide a run-time analysis for our new
inference algorithm.
The high-level likelihood estimation procedure is described in
Fig~\ref{fig:overview}. First, the user specifies a dynamical system
and a (multi-)~set of demonstrations. Then, using a user-defined
mechanism, a candidate task specification is selected. The
system then creates a compressed representation of the composition
of the dynamical system with the task specification. Then, in parallel,
the maximum causal entropy policy is estimated and the
demonstrations are themselves encoded as bit-vectors. Finally, the likelihood
of generating the encoded demonstrations is computed.

\begin{figure}[h]
  \centering
  \scalebox{.65}{ 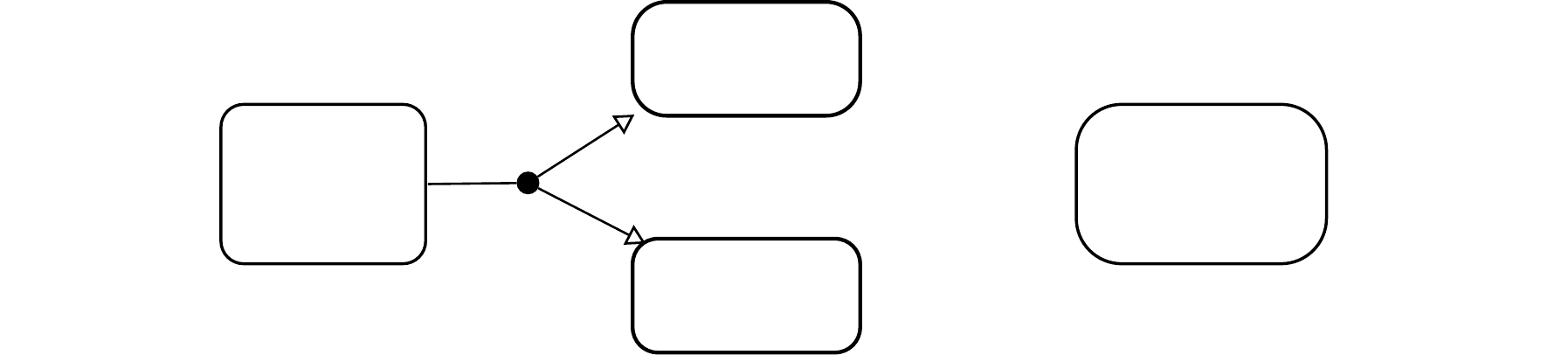 }
  \caption{
    High level likelihood estimation procedure described in this
    paper.\label{fig:overview}
  }
  \label{fig:overview}
\end{figure}

\vspace{-1em}
There are three computational bottlenecks in the
compressed scheme. First, given a candidate specification, $\varphi$,
one needs to construct $\mathcal{B}$. As argued in
Sec~\ref{sec:construct}, this takes time at most polynomial in the
horizon, monitoring automata size, and MDP size (in the random-bit
model). Second is the process of computing $Q$ and $V$ values by tuning
the rationality coefficient to match a particular satisfaction probability.
Just as with the na\"ive run-time~\eqref{runtime_naive}, this process takes time linear in the
size of $|\mathcal{B}|$ and logarithmic in the inverse tolerance $1/\epsilon$. Further, using Corr~\ref{corr:bdd_size}, we know that $|\mathcal{B}|$ is at most linear in horizon and quasi-linear in the
MDP size. Thus, the policy computation takes time polynomial in
the MDP size and logarithmic in the inverse tolerance. Finally,
as before, evaluating the likelihoods takes time linear in the number
of demonstrations and the horizon. However, we now require an additional
step of finding coin-flips which are consistent with the demonstrations.
Thus, the compressed run-time is bounded by:

\begin{equation}
  O\bigg(
    \Big(\hspace{-0.4em}
      \underbrace{|X|}_{\#\text{Demos}} \hspace{-0.4em}\cdot
      \hspace{-1em}
          \overbrace{\log\left (\epsilon^{-1}\right)}^{\text{Feature Matching}}
      \hspace{-1em}
    \Big)
    \cdot\text{POLY}\Big(
      \hspace{-0.5em}
        \overbrace{\tau}^{\text{Horizon}}
      \hspace{-1em}
      ,
      \hspace{-0.4em}
        \underbrace{|S|, |S_{|\varphi|}|, |A|}_{\text{Composed MDP size}}
      \hspace{-0.8em}
      ,
      \hspace{-3em}
        \overbrace{2^q}^{\#\text{Coin Flip Outcomes}}
      \hspace{-3em}
      \Big)\bigg)
\end{equation}

\begin{remark}
  In practice, this analysis is fairly conservative since BDD
  composition is often fast, the bound given by Corr~\ref{corr:bdd_size}
  is loose, and the SAT queries under-consideration are often trivial.
\end{remark}

\section{Additional Model Refinements}
\subsection{Conditioning on Valid Actions:}\label{sec:valid}

So far, we have assumed that the number of actions is a power of
$2$. Functionally, this assumption makes it so each assignment to the
action decision bits corresponds to a valid action. Of course, general
MDPs have non-power of $2$ action sets, and so it behooves us to adapt
our method for such settings. The simplest way to do so is to use a
3-terminal Binary Decision Diagram.  In particular, while
each decision is still Boolean, there has now three possible types of
leaves, $0$, $1$, and $\bot$. In the adapted algorithm, edges leading
to $\bot$ are simply ignored, as they semantically correspond to
invalid assignments to action or coin flip bits. A similar analysis
can be done using these three valued decision diagrams, and as with
BDDs, there exist efficient implementations of multi-terminal BDDs.

\begin{remark}
  This generalization also opens up the possibility of state dependent
  action sets, where $A$ is now the union of all possible actions,
  e.g, disable the action for moving to the right when the agent is on
  the right edge of the grid.
\end{remark}

\subsection{Choice of binary co-domain:}

One might wonder how sensitive this formulation is to the choice of
$R(\xi) = \theta \cdot \varphi(\xi)$. In particular, how does
changing the co-domain of $\varphi$ from $\{0, 1\}$ to any other real
values, i.e.,
\[\varphi': (A \times S)^\tau \to \{a, b\},\]
change the likelihood estimates in our maximum causal entropy model.
We briefly remark that, subject to some mild technical assumptions,
almost any two real values could be used for $\varphi$'s
co-domain. Namely, observe that unless both $a$ and $b$ are zero,
the expected satisfaction probability, $p$,  is in one-to-one correspondence with
the expected value of $\varphi'$, i.e.,
\[
\E[\varphi'] = a\cdot p + b\cdot (1 - p).
\]
Thus, if a policy is feature matching for $\varphi$, it must be
feature matching for $\varphi'$ (and vice-versa). Therefore, the space
of consistent policies is invariant under such
transformations. Finally, because the space of policies is unchanged,
the maximum causal entropy policies must remain unchanged.  In
practice, we prefer the use of $\{0, 1\}$ as the co-domain for
$\varphi$ since it often simplifies many calculations.

\subsection{Variable Episode Lengths (with discounting):}\label{sec:discount}

As earlier promised, we shall now discuss how to extend our model to
include variable length episodes. For simplicity, we shall limit our
discussion to the setting where at each time step, the probability
that the episode will end is $\gamma \in (0, 1]$.
As we previously discussed, this can be modeled by introducing a sink
state, $\$$, representing the end of an episode~\eqref{discount}. In
the random bit model, this simply adds a few additional enviroment
coin flips, corresponding to the enviroments new transitions to the
sink state.
\begin{remark}
Note that when unrolled, once the end of episode transition happens,
all decisions are assumed inconsequential w.r.t $\varphi$. Thus, all
subsequent decisions will be compressed by in the BDD, $\mathcal{B}$.
\end{remark}

\noindent
Finally, observe that the probability that the episode ending
increases exponentially, implying that the planning horizon need not
be too big, i.e., the probability that the episode has not
ended by timestep, $\tau \in \Nat$, is: $(1-\gamma)^{\tau}.$ Thus,
letting $\tau = \ceil{\ln(\nicefrac{\epsilon}{1 - \gamma})}$ ensures that with
probability at least $1 - \epsilon$ the episode has ended.

\section{Experiment}
\vspace{-0.7em}
Below we report empirical results that provide evidence that our
proposed technique is robust to demonstration errors and that the
produced BDDs are smaller than a na\"ive dynamic programming scheme.
To this end, we created a reference implementation~\cite{mce-code} in
Python. BDD and SAT solving capabilities are provided via
dd~\cite{dd} and pySAT~\cite{pysat} respectively. To encode the task
specifications and the random-bit model MDP, we leveraged the py-aiger
ecosystem~\cite{pyAiger} which includes libraries for modeling Markov
Decision Processes and encoding Past Tense Temporal Logic as
sequential circuits.

\begin{wrapfigure}[10]{r}{3.5cm}
  \vspace{-3em}
  \centering
  \scalebox{0.8}{ 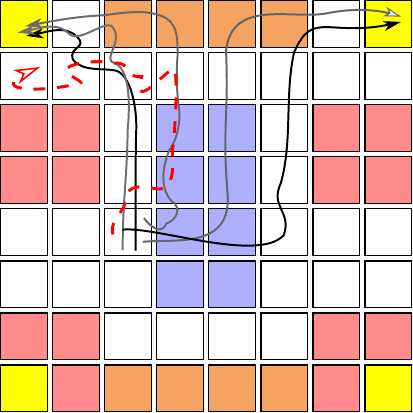 }
  \caption{\label{fig:demos}}
\end{wrapfigure}

\textbf{Problem: }Consider a gridworld where an agent can
attempt to move up, down, left, or right; however, with probability
$1/32$, the agent slips and moves left. Further, suppose a demonstrator
has provided the six unlabeled demonstrations shown in~\figref{demos}
for the task: ``Within 10 time steps, touch a yellow (recharge) tile
while avoiding red (lava) tiles. Additionally, if a blue (water) tile
is stepped on, the agent must step on a brown (drying) tile before
going to a yellow (recharge) tile.'' All of the solid paths
satisfy the task.  The dotted path fails because the agent
keeps slipping left and thus cannot dry off by $t=10$. Note
that due to slipping, all the demonstrations that did not enter the
water are sub-optimal.

\vspace{-0.6em}
\begin{center}
  \small
  
\begin{tabular}{lccc}\label{tbl:results}
Spec & Policy Size & ~~~~~ROBDD~~~~~ & Relative Log Likelihood  \\
     &  (\#nodes)  & build time      & (compared to True)     \\
\hline
true & 1           & 0.48s           & 0\\
\(\varphi_1\) = Avoid lava & 1797 & 1.5s & -22\\
\(\varphi_2\) = Eventually Recharge & 1628 & 1.2s & 5\\
\(\varphi_3\) = Don't recharge while wet & 850 & 1.6s & -10\\
\(\varphi_4 \eqdef \varphi_1 \wedge \varphi_2\) & 523 & 1.9s & 4\\
\(\varphi_5 \eqdef \varphi_1 \wedge \varphi_3\) & 1913 & 1.5s & -2\\
 \(\varphi_6 \eqdef \varphi_2 \wedge \varphi_3\) & 1842 & 2s & 15\\
 \(\varphi^* \eqdef \varphi_1 \wedge \varphi_2 \wedge \varphi_3\) & 577  & 1.6s & 27\\
\end{tabular}


\end{center}

\noindent
\textbf{Results:} For a small collection of specifications, we have
computed the size of the BDD, the time it took to construct the
BDD, and the \emph{relative} log likelihoods of the demonstrations\footnotemark,
\vspace{-0.7em}
\begin{equation}\label{eq:rel_log_prob}
  \text{RelativeLogLikelihood}(\varphi) \eqdef \ln\left(
      \frac{\Pr(\text{demos}~|~\varphi)}
           {\Pr(\text{demos}~|~\text{true})}
  \right),
\end{equation}

\vspace{-0.5em}
where each maximum entropy policy was fit to match the corresponding
specification's empirical satisfaction probability.
\footnotetext{The maximum entropy policy for $\varphi=\text{true}$ applies actions uniformly at random.}
We remark that the computed BDDs are small compared to other
straw-man approaches.  For example, an explicit construction of the
product of the monitor, dynamics, and 
the current time step would require space given by:
\vspace{-0.5em}
\begin{equation}\label{eq:n_nodes}
  \tau \cdot|S|\cdot |A|\cdot |S_\varphi| = (10\cdot 8\cdot 8 \cdot 4) \cdot |S_\varphi| = 2560 \cdot |S_\varphi|
  \vspace{-0.5em}
\end{equation}

The resulting BDDs are much smaller
than~\eqref{n_nodes} and the na\"ive unrolled decision tree.
We note that the likelihoods appear to (qualitatively)
match expectations. For example, \textbf{despite} an unlabeled
negative example, the demonstrated task, $\varphi^*$, is the most
likely specification. Moreover, under the second most likely
specification, which omits the avoid lava constraint, the sub-optimal
traces that do not enter the water appear more attractive.

Finally, to emphasize the need for our causal extension, we compute
the likelihoods of
$\varphi^*, \varphi_1, \varphi_2$ for our opening
example (\figref{enterlava2}) using both our causal model and the prior non-causal model~\cite{max_ent_spec_learning}. Concretely, we take $\tau = 15$, a slip
probability of 1/32, and fix the expected satisfaction probability to
0.9. The trace shown in~\figref{enterlava2} acts as the sole (failed)
demonstration for $\varphi^*$.  As desired, our causal extension
assigned more than 3 times the relative likelihood to $\varphi^*$
compared to $\varphi_1$, $\varphi_2$, and $\true$. By contrast, the
non-causal model assigns relative log likelihoods $(-2.83, -3.16,
-3.17)$ for $(\varphi_1, \varphi_2, \varphi^*)$. This implies that (i)
$\varphi^*$ is the least likely specification and (ii) each
specification is less likely than $\true$!

\vspace{-1.3em}
\section{Conclusion and Future Work}\label{sec:conclusion}
\vspace{-0.5em}
Motivated by the problem of learning specifications from
demonstrations, we have adapted the principle of maximum causal
entropy to provide a posterior probability to a candidate task
specification given a multi-set of demonstrations. Further, to exploit
the structure of task specifications, we proposed an algorithm that
computes this likelihood by first encoding the unrolled Markov
Decision Process as a reduced ordered binary decision diagram (BDD).
As illustrated on a few toy examples, BDDs are often much smaller
than the unrolled Markov Decision Process and thus could enable
efficient computation of maximum causal entropy likelihoods, at least
for well behaved dynamics and specifications.

Nevertheless, two major questions remain unaddressed by this
work. First is the question of how to select which specifications to
compute likelihoods for. For example, is there a way to systematically
mutate a specification to make it more likely and/or is it possible to
systematically reuse computations for previously evaluated
specifications to propose new specifications.

Second is how to set prior probabilities. Although we have
largely ignored this question, we view the problem of setting good
prior probabilities as essential to avoid over fitting and/or making
this technique require only one or two demonstrations. However, we
note that prior probabilities can make inference arbitrarily more
difficult since any structure useful for optimization imposed by our
likelihood estimate can be overpowered.

Finally, additional future work includes extending the formalism to
infinite horizon specifications, continuous dynamics, and
characterizing the optimal set of teacher demonstrations.


%
%
%

{\vspace{0.5em} \footnotesize
  \noindent\textbf{Acknowledgments}:
  We would like to thank the anonymous referees as well as Daniel
  Fremont, Ben Caulfield, Marissa Ramirez de Chanlatte, Gil Lederman,
  Dexter Scobee, and Hazem Torfah for their useful suggestions and
  feedback.  This work was supported in part by NSF grants 1545126
  (VeHICaL) and 1837132, the DARPA BRASS program under agreement
  number FA8750-16-C0043, the DARPA Assured Autonomy program, Toyota
  under the iCyPhy center, and Berkeley Deep Drive.  }

%
%

\raggedright
\bibliographystyle{splncs04}
%
\bibliography{refs}

\end{document}